\documentclass[letterpaper, 10 pt, conference]{ieeeconf}  
\usepackage{cite}
\usepackage{soul}
\usepackage{tikz}
\usetikzlibrary{arrows.meta, positioning, fit}

\usepackage{graphicx}   
\usepackage{booktabs}   
\usepackage{multirow}   
\usepackage[section]{placeins} 
\usepackage{cuted}      
\usepackage{caption}    
\usepackage{booktabs}
\usepackage{tabularx}
\usepackage{placeins}
\usepackage{siunitx}
\usepackage{booktabs}
\usepackage{siunitx}
\usepackage{xurl}                 
\usepackage[hidelinks]{hyperref}  
\usepackage{xcolor}
\usepackage{soul}
\usepackage{xcolor}

\IEEEoverridecommandlockouts                              

\usepackage{amsmath} 
\usepackage{amssymb}  

\title{\LARGE \bf
Robust Building Damage Detection in Cross-Disaster Settings Using Domain Adaptation

}

\author{%
\makebox[\textwidth][c]{%
\begin{tabular}{@{}p{0.48\textwidth}@{\hspace{0.04\textwidth}}p{0.48\textwidth}@{}}
\centering
{\large Asmae Mouradi}\\
\textit{School of Computing}\\
Wichita State University\\
Wichita, Kansas, USA\\
axmouradi@shockers.wichita.edu
&
\centering
{\large Shruti Kshirsagar}\\
\textit{School of Computing}\\
Wichita State University\\
Wichita, Kansas, USA\\
shruti.kshirsagar@wichita.edu
\end{tabular}%
}%
}%

\begin{document}

\maketitle
\thispagestyle{empty}
\pagestyle{empty}

\begin{abstract}
Rapid structural damage assessment from remote sensing imagery is essential for timely disaster response. Within human–machine systems (HMS) for disaster management, automated damage detection provides decision-makers with actionable situational awareness. However, models trained on multi-disaster benchmarks often underperform in unseen geographic regions due to domain shift—a distributional mismatch between training and deployment data that undermines human trust in automated assessments. We explore a two-stage ensemble approach using supervised domain adaptation (SDA) for building damage classification across four severity classes. The pipeline adapts the xView2 first-place method to the Ida-BD dataset using SDA and systematically investigates the effect of individual augmentation components on classification performance. Comprehensive ablation experiments on the unseen Ida-BD test split demonstrate that SDA is indispensable: removing it causes damage detection to fail entirely. Our pipeline achieves the most robust performance using SDA with unsharp enhanced RGB input, attaining a Macro F1 of 0.5552. These results underscore the critical role of domain adaptation in building trustworthy automated damage assessment modules for HMS-integrated disaster response.
\end{abstract}




\section{Introduction}

Human machine systems (HMS) combine human decision-making with automated tools to aid disaster response and management. Climate change has increased disaster frequency and intensity, leading to increased demands on these systems\cite{ipcc2012}. HMS frameworks assist emergency responders in assessing situations and manage resources during hurricanes, earthquakes, and floods. In HMS, the proposed model can serve as a decision-support tool that helps analysts focus on the most severely affected regions while reducing manual screening effort. In the past, satellite-based building damage assessment relied on manual comparison of pre- and post-disaster imagery, leading to a significant response time bottleneck \cite{dong2013}. Automation is needed to inspect large regions with numerous damage levels, which takes time. Although remote sensing imagery is commonly utilized for wildfire detection~\cite{filipponi2019}, land-use analysis~\cite{foody2003}, and disaster management~\cite{schumann2018}, the human role in analyzing this data has not altered significantly. Automating structural damage detection speeds up and scales disaster response, allowing human operators to focus on higher-level decision-making and supervision activities in HMS processes. 

Deep learning has streamlined the building damage detection task and generated standardized damage mapping solutions for human decision-makers quickly. Encoder-decoder architectures such as U-Net~\cite{ronneberger2015} have been widely adopted for pixel-wise semantic labeling of building footprints and damaged regions. Prior work has explored both post-disaster-only approaches such as support vector machine-based methods applied to high-resolution QuickBird imagery~\cite{li2009} and combined pre- and post-disaster strategies that leverage texture based and shape based features to improve assessment accuracy~\cite{cooner2016}. More recent advances include attention mechanisms for capturing long-range spatial dependencies~\cite{hao2021, kshirsagar2022affective}, cross-directional feature fusion networks~\cite{shen2021bdanet}, and Siamese architectures for change detection~\cite{khvedchenya2021siamese}. However, a fundamental challenge arises when these models trained on multi-disaster benchmark datasets are deployed in unseen geographic regions or disaster scenarios not represented in the training distribution \cite{parupati2024enhancing, kshirsagar2026geographic}. This phenomenon is known as \textit{domain shift}~\cite{hertel2025}. Domain shift causes significant degradation in classification accuracy and consistency. Variations in imaging sensors, geographic environments, atmospheric conditions, and catastrophic event types create a distributional misalignment between training and inference data. Using domain adaptation (DA) approaches, models can transfer learned information across diverse data distributions \cite{hafner2022}. DA techniques can be supervised (annotations in both source and target domains), unsupervised (labels exclusively in source domain), or semi-supervised (limited target labels and unlabeled data) based on label availability \cite{tuia2016}. Selecting unlabeled source domains for remote sensing adaptation has been studied \cite{geiss2022}. In building damage assessment, Parupati et al. \cite{parupati2025,kshirsagar2026geographic} found that supervised fine-tuning on target-domain data enhances generalization compared to pretrained-only and unsupervised DA baselines. To promote research in automated post-event damage evaluation, the xView2 Challenge \cite{gerard2024} created a competitive benchmark on the xBD dataset\cite{gupta2019xbd}. The top-ranked solutions \cite{durnov2020,seferbekov2020,khvedchenya2020third} in the challenge represent the current state-of-the-art methods. Augmentation methods along with domain adaptation have been widely utilized to improve model generalizability~\cite{K1, K2, K3, parupati2025, kshirsagar2026geographic}. However, the robustness of these algorithms and the effect of different augmentation components on overall performance remain largely unexplored.

In this work, we proposed to use a two-stage ensemble supervised domain adaptation strategy using transfer learning from the xBD source dataset~\cite{gupta2019xbd} to the Ida-BD target dataset~\cite{lee2022idabd}.

In summary, our contributions are as follows:
\begin{itemize}

    \item We explore a two-stage ensemble pipeline combining building segmentation (Stage~1) and damage classification (Stage~2) with fusion augmentation techniques, achieving state-of-the-art performance across multiple damage severity classes on Ida-BD, surpassing the results reported in~\cite{parupati2025}.
    \item We perform ablation experiments with domain adaptation and fusion augmentation methods. These insights provide actionable guidance for designing reliable automated damage assessment modules within broader human-machine disaster management systems.
\end{itemize}

The remainder of this paper is organized as follows. Section~\ref{sec:method} details the proposed two-stage method. Section~\ref{sec:setup} presents the experimental setup. Section~\ref{sec:results} discusses the results and ablation analysis. Finally, Section~\ref{sec:conclusion} concludes the paper.

\section{Proposed Method}

\begin{figure*}[t]
  \centering
  \includegraphics[width=0.8\textwidth]{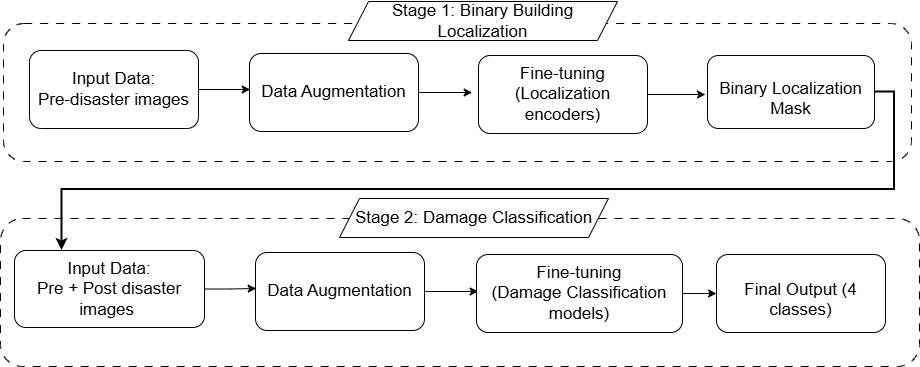}
  \caption{Overview of the proposed two-stage pipeline for cross-disaster damage assessment. Stage 1 performs building localization to generate a building mask. Stage 2 uses pre- and post-disaster images, for four-class damage classification.}
  \label{fig:pipeline}
\end{figure*}

\label{sec:method}
The proposed method is based on a two-stage end-to-end pipeline from pre-event and post-event satellite imagery. Here, we explore the benefits of implementing a two-stage pipeline. In this section, we start by describing the steps followed to obtain the two-stage pipeline, then describe the data augmentation approach.
\subsection {Two-stage classification approach }
The two-stage classification approach has been described in detail in Figure \ref{fig:pipeline}
This figure shows the framework's two-stage training pipeline. The pipeline uses pre-disaster satellite imagery in Stage 1 (Binary Building Localization) to increase training variations. The 12 localization encoder models, pretrained on the xBD dataset using the xView2 first-place solution, are fine-tuned using the augmented images. The supervised domain adaptation stage of fine-tuning aligns models to the Ida-BD target domain. Stage 1 generates a binary localization mask that distinguishes images as building or background. The connecting arrow shows that this mask feeds into Stage 2 (Damage Classification). Stage 2 generates the six-channel input representation from pre- and post-disaster images. Fusion augmentation—edge detection, contrast enhancement (CLAHE), and unsharp masking—prepares these images for damage classification models, which are fine-tuned from xBD-pretrained weights onto Ida-BD. Particularly, Stage 1's binary localization mask gates Stage 2, assuring damage categorization only over building pixels. A four-class damage prediction map is generated, labeling building images as No-Damage, Minor, Major, or Destroyed.

This proposed two-stage ensemble approach shows how sequential gating decreases background false positives and improves damage estimates for human operators.
Transfer learning is used to initialize models with xBD-pretrained weights (source domain) and fine-tune them on annotated Ida-BD data (target domain). This technique immediately overcomes the sensor, geographic, and disaster-type domain gap between xBD and Ida-BD. We compare this to a no-DA baseline where pretrained xBD checkpoints are applied straight to Ida-BD without fine-tuning to measure how domain adaptation affects generalization performance. This comparison directly analyzes how much adaptation is needed to provide outputs that human operators can trust for unique disaster scenarios from an HMS perspective.

\subsection{Augmentation Methods}

Post-disaster satellite imagery can be highly variable due to atmospheric conditions, sensor angle variations, and debris-induced texture changes. We study several types of image-level augmentation strategies, employed singly and in combination, which increase complementary visual cues essential to damage assessment to improve model robustness against these appearance differences. Based on the fusion augmentation strategy described in \cite{ghaffar2019data}, these methods are explained below.

\subsubsection{Edge Detection}

Edge detection utilizes visualization gradients to emphasize satellite imagery's structural boundaries. Edges indicate roof, wall, and debris field boundaries that may indicate structural damage. Injecting edge cues into the model input helps the network detect transitions between intact and damaged regions. An edge map $E$ is a single-channel feature map that emphasizes object boundaries.

\subsubsection{Contrast Enhancement (CLAHE)}

Contrast Limited Adaptive Histogram Equalization (CLAHE) increases local contrast and highlights subtle damage-related patterns. CLAHE is better for satellite imaging with variable lighting conditions than global histogram equalization because it limits contrast amplification on local image tiles to avoid noise enhancement. Contrast-enhanced image $C$ shows low-contrast damage signatures, including discolored roofs, wet regions, and partially covered debris that may be missed in raw RGB. This improvement helps distinguish No-Damage buildings from ones with minimal surface damage.

\subsubsection{Unsharp Masking}

Unsharp masking enhances high-frequency details by subtracting a Gaussian-blurred image and reinjecting the amplified residual back into the original:
\begin{equation}
U = I + \lambda\,(I - G_\sigma * I),
\label{eq:unsharp}
\end{equation}
where $G_\sigma * I$ denotes the Gaussian-blurred image with kernel width $\sigma$, and $\lambda$ controls the sharpening strength. This operation accentuates fine-grained structural details including cracks, splintered roofing materials, scattered debris fields, and deformed structural elements that are characteristic of severe damage categories such as Major and Destroyed. Among the augmentation components investigated in this work, unsharp masking proves most effective for recognizing heavily damaged structures, as it amplifies the high-frequency textural disruptions that distinguish destroyed buildings from those with only minor damage.




\subsubsection{Fusion Augmentation}
The fusion augmentation strategy combines all three enhancement methods with the original image through a weighted linear combination:
\begin{equation}
I_{\text{fuse}} = \alpha\, I + \beta\, E + \gamma\, C + \delta\, U,
\label{eq:fusion_mat}
\end{equation}
where the weights $(\alpha, \beta, \gamma, \delta)$ control the relative contribution of each component and typically satisfy $\alpha + \beta + \gamma + \delta = 1$. In this work, these weights are chosen as fixed settings in the augmentation design and remain the same during training and evaluation. The fusion output integrates boundary cues (edges), locally enhanced contrast (CLAHE), and sharpened fine details (unsharp masking) into a single enriched representation. However, as our ablation experiments reveal in Section~\ref{sec:results}, combining all components simultaneously does not always yield the best performance; the interactions between enhancement methods can introduce conflicting cues that degrade the classification of minority damage classes. This motivates our systematic evaluation of individual components and pairwise combinations alongside the full fusion method.

Each configuration is evaluated both with  supervised domain adaptation to isolate the individual and combined effects of augmentation and SDA on damage classification performance.

\section{Experimental Setup}
\label{sec:setup}
In this section, we describe the datasets, benchmark system, and performance metric used in these experiments.
\subsection{Datasets}
\subsubsection{xBD Dataset (Source Domain)}

The xBD dataset~\cite{gupta2019xbd} is a large-scale remote sensing imagery dataset for structural damage assessment resulting from various natural disasters, including hurricanes, tornadoes, wildfires, earthquakes, flooding events, and volcanic events. The dataset comprises very-high-resolution image pairs (organized as tiles of size $1024 \times 1024$ pixels) captured in pre-event and post-event conditions for 19 natural disaster, totaling 850,736  building annotations recorded across several geographic regions. The xBD dataset involves two analytical tasks: (1) building localization and (2) damage classification across four severity classes. The dataset is divided into four subsets: train, tier3, test, and holdout. The training, testing, and validation sets include data corresponding to identical disaster occurrences, while the tier3 set includes data related to supplementary disaster events not represented in the remaining subsets. In our framework, we utilize xBD exclusively as the source domain for initializing model weights.

\subsubsection{Ida-BD Dataset (Target Domain)}

\begin{figure}[t]
  \centering
  \includegraphics[width=\columnwidth]{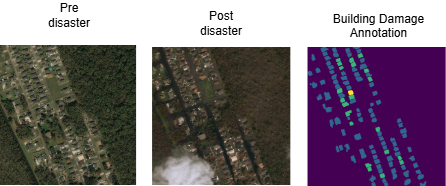}
  \caption{Representative sample from the Ida-BD dataset showing the pre-disaster image, post-disaster image, and ground-truth building damage annotation mask.}
  \label{fig:idabd_example}
\end{figure}

Ida-BD dataset \cite{lee2022idabd} includes 87 pre- and post-disaster image pairs from the WorldView-2 satellite, captured in Louisiana, USA, before and after Hurricane Ida in 2021. Unlike most xBD events, this dataset has a discrete geographic region, disaster kind, and imaging sensor, making it an ideal target domain for cross-disaster generalization. Our supervised adaptation technique addresses the domain change caused by sensor features, regional building types, vegetation patterns, and hurricane-specific damage signatures between xBD and Ida-BD. We divide the Ida-BD dataset into training (80\%), validation (10\%), and test (10\%) splits for our studies. The training split fine-tunes, the validation split selects models and optimizes model parameters, and the held-out test split evaluates the pipeline. Figure~\ref{fig:idabd_example} illustrates a representative sample from the Ida-BD dataset~\cite{lee2022idabd}, showing the pre-disaster image, the corresponding post-disaster image, and the ground-truth building damage annotation mask. The annotation mask color-codes each building according to its assigned damage severity level, providing pixel-level supervision for both localization and classification tasks.
Pre-disaster and post-disaster images were converted to RGB, scaled to the range $[0,1]$, and normalized using ImageNet mean and standard deviation. The pre-disaster and post-disaster RGB images were then concatenated to form a 6-channel input tensor.



\subsection{Benchmark System: xView2 Challenge}

The xView2 competition \cite{gerard2024} is a competitive benchmark based on the xBD dataset designed to promote research in automatic post-event damage evaluation using remote sensing imagery. The three winning solutions \cite{durnov2020,seferbekov2020,khvedchenya2020third} as described in Table~\ref{tab:xview2} represent the current state of the art in automated building damage assessment from satellite imagery. We also used a benchmark from \cite{parupati2025}, where they utilized the same cross-data setting.
 Table~\ref{tab:xview2} presents the F1 scores of the top three solutions on the official xBD test set. Our pipeline initializes from the first-place solution's pretrained weights, which employ an ensemble of four U-Net-based architectures, including DPN92, ResNet34, SE-ResNeXt50, and SENet154 backbones, each trained with three distinct random seeds, resulting in 12 models altogether.

\begin{table}[t]
\centering
\caption{F1 scores of the top-three xView2 Challenge solutions evaluated on the xBD test set.}
\label{tab:xview2}
\resizebox{\columnwidth}{!}{%
\begin{tabular}{lccccc}
\toprule
\textbf{Method} & \textbf{Loc.} & \textbf{No-Dmg} & \textbf{Minor} & \textbf{Major} & \textbf{Dest.} \\
\midrule
1st (ResNet34 + SE-ResNeXt50)~\cite{durnov2020} & 0.862 & 0.915 & 0.639 & 0.782 & 0.854 \\
2nd (DPN92 + DenseNet161)~\cite{seferbekov2020} & 0.853 & 0.902 & 0.618 & 0.770 & 0.849 \\
3rd (ResNet + DenseNet + EffNet)~\cite{khvedchenya2020third} & 0.847 & 0.907 & 0.617 & 0.765 & 0.846 \\
\bottomrule
\end{tabular}%
}
\end{table}

\subsection{Performance Measures}
\label{sec:metrics}

We evaluate the proposed framework using the F1 score, defined as the harmonic mean of precision and recall for each class $c$:
\begin{equation}
F1_c = \frac{2 \cdot \text{Precision}_c \cdot \text{Recall}_c}{\text{Precision}_c + \text{Recall}_c}.
\label{eq:f1}
\end{equation}

\textbf{Macro-F1 score:} The macro-average across the four damage classes, computed as : 

\begin{equation}
\text{Macro-F1} = \frac{1}{4} \sum_{c=1}^{4} F1_c,
\label{eq:macro_f1}
\end{equation}
where $c$ indexes the damage classes $c \in \{1, 2, 3, 4\}$. The Macro-F1 provides a single summary metric for HMS-integrated disaster response that shows whether the automated system makes balanced, trustworthy predictions across the damage spectrum.

\section{Results and Discussion}
\label{sec:results}
In this section, we present the experimental results for building damage detection in a cross-data setting  and discuss our findings.

\subsection{Effect of supervised domain adaptation }

\begin{figure}[t]
  \centering
  \includegraphics[width=\columnwidth]{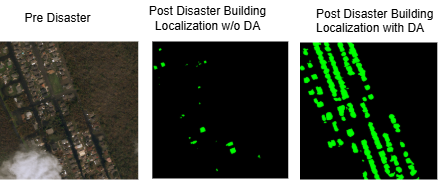}
  \caption{Qualitative comparison of Stage-1 building localization on the Ida-BD target domain: (a) input image, (b) localization result without domain adaptation, and (c) localization result with supervised domain adaptation.}
  \label{fig:stage1_da_comparison}
\end{figure}

\begin{table*}[t]
\centering
\caption{Effect of supervised domain adaptation (SDA) on building damage detection across the Ida-BD test set (F1 scores). Bold values indicate best performance per column.}
\label{tab:ablation1}
\resizebox{\textwidth}{!}{%
\begin{tabular}{lcccccc}
\toprule
\textbf{Methods} & \textbf{Localization} & \textbf{No-Damage} & \textbf{Minor} & \textbf{Major} & \textbf{Destroyed} & \textbf{Macro F1} \\
\midrule
Two-Stage Ensemble + RGB only (baseline) + DA & 0.8494 & 0.7108 & 0.4463 & \textbf{0.5420} & \textbf{0.5073} & \textbf{0.5516} \\
Two-Stage Ensemble + Fusion Aug. + DA & 0.8493 & 0.7312 & 0.4508 & 0.5271 & 0.3877 & 0.5242 \\
Two-Stage Ensemble + w/o Fusion Aug. + DA & 0.8493 & \textbf{0.7336} & \textbf{0.4671} & 0.5302 & 0.2321 & 0.4908 \\
Two-Stage Ensemble + Fusion Aug. + w/o DA & \textbf{0.8661} & 0.6424 & 0.0039 & 0.0480 & 0.0000 & 0.1736 \\

\bottomrule
\end{tabular}%
}
\end{table*}
\begin{table*}[t]
\centering
\caption{Ablation study of individual and combined augmentation components with SDA on the Ida-BD test set (F1 scores). Bold values indicate best performance per column}
\label{tab:ablation}
\resizebox{\textwidth}{!}{%
\begin{tabular}{lcccccc}
\toprule
\textbf{Methods} & \textbf{Localization} & \textbf{No-Damage} & \textbf{Minor} & \textbf{Major} & \textbf{Destroyed} & \textbf{Macro F1} \\
\midrule

Two-Stage Ensemble + RGB + Contrast + DA & 0.8494 & 0.7334 & 0.4475 & 0.5374 & 0.0412 & 0.4399 \\
Two-Stage Ensemble + RGB + Unsharp + DA & 0.8494 & 0.7182 & 0.4313 & 0.5558 & \textbf{0.5156} & \textbf{0.5552} \\
Two-Stage Ensemble + RGB + Edges + DA & 0.8494 & 0.7225 & 0.4309 & 0.5541 & 0.4362 & 0.5359 \\
Two-Stage Ensemble + RGB + Contrast + Edges + DA & \textbf{0.8877} & \textbf{0.7338} & 0.4236 & 0.5231 & 0.1957 & 0.4690 \\
Two-Stage Ensemble + RGB + Unsharp + Edges + DA & 0.8789 & 0.7011 & 0.4261 & 0.5161 & 0.4652 & 0.5271 \\
Two-Stage Ensemble + RGB + Unsharp + Contrast + DA & 0.8867 & 0.7304 & 0.4279 & \textbf{0.5602} & 0.3653 & 0.5210 \\
Two-Stage Ensemble + Fusion Aug. + DA & 0.8493 & 0.7312 & \textbf{0.4508} & 0.5271 & 0.3877 & 0.5242 \\
Two-Stage Ensemble + RGB + Fusion Aug. + DA & 0.8494 & 0.7160 & 0.4310 & 0.4875 & 0.0000 & 0.4086 \\
\bottomrule
\end{tabular}%
}
\end{table*}

\begin{table}[t]
\centering
\caption{Comparison with benchmark systems on the Ida-BD dataset in terms of F1 score.}
\label{tab:comparison}
\resizebox{\columnwidth}{!}{%
\begin{tabular}{lccccc}
\toprule
\textbf{Method} & \textbf{Loc.} & \textbf{No-Dmg} & \textbf{Minor} & \textbf{Major} & \textbf{Dest.} \\
\midrule
Pretrained only~\cite{parupati2025} & 0.806 & 0.667 & 0.211 & 0.154 & 0.041 \\
+ Augmentation~\cite{parupati2025} & 0.815 & 0.663 & 0.235 & 0.173 & 0.052 \\
+ Supervised DA~\cite{parupati2025} & 0.842 & 0.672 & 0.292 & 0.184 & 0.095 \\
+ Supervised DA + Aug.~\cite{parupati2025} & 0.849 & 0.696 & 0.315 & 0.192 & 0.117 \\
+ Unsupervised DA-CORAL~\cite{parupati2025} & 0.815 & 0.658 & 0.275 & 0.135 & 0.059 \\
\midrule
\textbf{Ours (RGB + Unsharp + DA)} & \textbf{0.849} & \textbf{0.718} & \textbf{0.431} & \textbf{0.556} & \textbf{0.516} \\
\textbf{Ours (best per-class)} & \textbf{0.888} & \textbf{0.734} & \textbf{0.467} & \textbf{0.560} & \textbf{0.516} \\
\bottomrule
\end{tabular}%
}
\end{table}

In our first experiment, we investigate the effect of supervised domain adaptation (SDA) in a cross-data setting. The model is fine-tuned using the two-stage ensemble approach on the xBD dataset before being trained on the Ida-BD dataset and reporting F1 scores on the unseen Ida-BD test split. We examine the fusion augmentation approach with SDA, with results in Table~\ref{tab:ablation1}.

Using fusion augmentation without supervised domain adaptation results in reasonable localization F1 (0.8661) but a significant drop in the other three classifications: Minor F1 drops to 0.0039, Major to 0.0480, and Destroyed to 0.0000, resulting in a damage Macro-F1 of only 0.1736. Differences in image sensors, geographic contexts, building designs, and hurricane-specific damage signatures make the domain gap between xBD and Ida-BD too significant for xView2-pretrained models to overcome without target-domain fine-tuning. To further illustrate the impact of SDA, Figure \ref{fig:stage1_da_comparison} exhibits a qualitative comparison of Stage-1 building localization on a post-disaster image. Without domain adaptation, the localization model makes inaccurate and noisy building area projections. SDA enhances localization accuracy, creating cleaner, more comprehensive building masks that match real structures. The illustrated comparison confirms the quantitative findings in Table~\ref{tab:ablation1}, proving the need for domain adaptation for accurate building detection across disasters. This result is particularly important for HMS: models without domain adaptation fail to identify severely damaged buildings and misclassify most minor and major damage, making them unreliable for human decision-making in unseen disaster scenarios.

\subsection{Ablation study: Effect of Individual Augmentation Components}

\begin{figure}[t]
  \centering
  \includegraphics[width=\columnwidth]{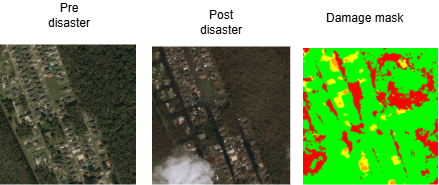}
    \caption{Qualitative damage detection result using RGB + Unsharp + SDA: (a) pre-disaster image, (b) post-disaster image, and (c) predicted damage mask. The predicted mask distinguishes the four damage levels no damage (green), minor (yellow), major (orange), and destroyed (red)}
  \label{fig:rgb_unsharp_da_damage}
\end{figure}

In this experiment, we aim to determine which element of fusion augmentation and SDA is more important for damage detection. Table~\ref{tab:ablation} reports the performance of each augmentation component combined with SDA.

Table~\ref{tab:ablation} shows that unsharp masking + DA yields the highest Destroyed F1 ($0.5156$) and total damage Macro-F1 ($\mathbf{0.5552}$), slightly beating the RGB-only baseline ($0.5516$). Sharpening highlights fine-grained structural characteristics, including cracks, debris edges, and roof fragmentation, that indicate serious damage. This configuration is best for HMS-integrated deployment, where accurate severity level detection is needed because of its balanced and robust performance across the damage spectrum. Figure~\ref{fig:rgb_unsharp_da_damage} displays the qualitative result of our top-performing combination (RGB + Unsharp + DA). The predicted damage mask shows that the model distinguishes between the four severity classes—no damage (green), minor (yellow), major (orange), and destroyed (red)—producing geographically consistent predictions that closely match the post-disaster damage distribution. This shows how unsharp masking and supervised domain adaptation allow the model to capture fine-grained structural damage cues for reliable HMS-integrated evaluation.
Next, Contrast enhancement + DA improves No-Damage classification ($0.7334$) but considerably decreases Destroyed performance ($0.0412$). Strong contrast changes may enhance dominating class features while masking severely harmed structures' tiny, scattered damage indications.  In operational HMS settings, this class-dependent sensitivity could have life-safety consequences, as failing to detect severely damaged buildings directly impacts rescue prioritization.
Interestingly, Edge detection + DA achieves moderate Destroyed F1 ($0.4362$) and competitive Macro-F1 ($0.5359$), demonstrating that boundary information helps damage assessment but cannot match unsharp masking for severe damage recognition. Furthermore, Contrast + Edges + DA results in the greatest localization F1 ($\mathbf{0.8877}$) and No-Damage F1 ($\mathbf{0.7338}$), but the damage Macro-F1 lowers to $0.4690$ due to lacking Destroyed detection ($0.1957$). In light of their varied optimal feature requirements, building detection and damage recognition do not necessarily improve each other. Similarly, Unsharp + Contrast + DA achieves the highest Major-Damage F1 ($\mathbf{0.5602}$) and strong localization $0.8867$, but Destroyed F1 decreases to $0.3653$, likely due to the contrast component counteracting the benefit of unsharp masking for severe damage. Finally, full fusion augmentation (all components) + DA shows that stacking all augmentation channels is ineffective. The combination appears to introduce conflicting feature cues that confuse the classifier on minority classes, effectively drowning out the damage-specific signals that individual components successfully capture.  


\subsection{Comparison with benchmark systems}

Table~\ref{tab:comparison} compares our results against benchmark systems described in \cite{parupati2025} since identical pretrained models and fusion augmentation approaches were utilized on Ida-BD. Our two-stage ensemble pipeline with supervised DA improves F1 scores across all damage categories. The most significant gains are seen in Major-Damage from $0.192$ to $0.556$, a 190\% increase and Destroyed from $0.117$ to $0.516$, a 341\% increase. Our ensemble strategy (averaging 12 models), optimized probability thresholding, building-pixel-restricted loss, and destroyed-aware crop sampling improve balance and reliability, especially for rare but operationally critical damage categories.


\section{Conclusion}
\label{sec:conclusion}

In this study, we apply data augmentation and domain adaptation to building damage detection from satellite imagery, focusing on cross-disaster generalization under domain shift for the building damage detection task. Our experimental results show that domain adaptation is the most crucial factor in damage detection in a cross-data setting. Unsharp-enhanced RGB input with supervised DA performs best, with a Macro-F1 of 0.5552. These findings support domain-adaptive techniques for integrating automated damage assessment into practical disaster response processes, as accurate model outputs are necessary for human-machine teaming. Future research will explore semi-supervised and unsupervised domain adaptation strategies to reduce annotation requirements and ensure the dependability of HMS-integrated deployment outputs. The code to reproduce our experiments is publicly available at \url{ https://github.com/asmaemou/xview2\_1st\_place\_solution}.

\section*{Acknowledgment}
The authors acknowledge funding support for this work from Kansas NSF EPSCoR First award grant and the Research and  Education Innovation grant.

\bibliographystyle{IEEEtran}
\bibliography{refs}

\end{document}